\documentclass[letterpaper, 10 pt, conference]{ieeeconf}  %

\IEEEoverridecommandlockouts                              %

\makeatletter
\def\endthebibliography{%
  \def\@noitemerr{\@latex@warning{Empty `thebibliography' environment}}%
  \endlist
}
\makeatother

\usepackage{cite}
\usepackage{graphicx} %
\usepackage[utf8]{inputenc}
\usepackage{amsmath} %
\usepackage{amssymb}  %
\usepackage{array}  %
\usepackage[usenames,dvipsnames]{xcolor}  %
\usepackage{hyperref}
\hypersetup{
    colorlinks,
    linkcolor = BlueViolet,
    urlcolor  = Blue,
    citecolor = Black,
    anchorcolor = Black
}
\usepackage[nohyperlinks,printonlyused,nolist]{acronym}
\usepackage{subcaption}
\usepackage{tikz}
\usepackage{pgfplots}
\pgfplotsset{compat=newest}
\usepgfplotslibrary{groupplots}
\usepackage{setspace}

\newcommand{\set}[1]{\bigl\{#1\bigr\}}
\newcommand{\tuple}[1]{\left(#1\right)}

\newcommand{\norm}[1]{\left\lVert#1\right\rVert}
\newcommand{\pnorm}[2]{\norm{#2}_{#1}}
\newcommand{\twonorm}[1]{\pnorm{2}{#1}}

\newcommand{\Reals}{\mathbb R}

\renewcommand{\v}{\mathbf{v}}
\newcommand{\x}{\mathbf{x}}

\newcommand{\f}{\mathbf{f}}

\renewcommand{\P}{\mathbf{P}}

\newcommand{\trans}[1]{#1^\top}

\usepackage{booktabs}

\newcommand*\centremathcell[1]{\omit\hfil$\displaystyle#1$\hfil\ignorespaces}

\DeclareMathOperator*{\argmin}{arg\,min}
\DeclareMathOperator*{\minimize}{minimize}
\DeclareMathOperator*{\subjectto}{subject\ to}

\newcommand{\bbm}{\begin{bmatrix}}
\newcommand{\ebm}{\end{bmatrix}}
\DeclareMathAlphabet{\mbf}{OT1}{ptm}{b}{n}

\newcommand{\state}{\x}
\newcommand{\humstate}{\tilde{\mathbf{x}}}
\newcommand{\init}{\text{o}}
\newcommand{\stateinit}{{\state_{\init}}}
\newcommand{\sspace}{\mathcal{X}}
\newcommand{\control}{\mathbf{u}}

\newcommand{\cspace}{\Reals^2}
\newcommand{\action}{\control}

\newcommand{\deltat}{{\delta~t~}}

\newcommand{\at}[2]{{#1}_{#2}}
\newcommand{\kpone}{{t+1}}
\newcommand{\kmone}{{t-1}}
\newcommand{\stateat}[1]{\at{\state}{#1}}
\newcommand{\robstateat}[1]{\id{r}{\at{\mathbf{x}}{#1}}}
\newcommand{\humstateat}[1]{\at{\humstate}{#1}}
\newcommand{\actionat}[1]{\at{\action}{#1}}
\newcommand{\controlat}[1]{\at{\control}{#1}}

\newcommand{\velvec}{\v}
\newcommand{\vel}{\velvec}

\newcommand{\xpos}{x}
\newcommand{\ypos}{y}
\newcommand{\head}{{\theta}}
\newcommand{\lvel}{{v}}
\newcommand{\dist}{{d}}

\newcommand{\avel}{{\omega}}

\newcommand{\robidmarker}{r}
\newcommand{\idRob}{\robidmarker}

\newcommand{\id}[2]{{#2^{(#1)}}}
\newcommand{\rob}[1]{\id{\robidmarker}{#1}}

\newcommand{\pref}{\text{intent}}

\newcommand{\robdyn}{\f}
\newcommand{\humdyn}{\mathbf{h}}

\newcommand{\numhumans}{N}
\newcommand{\numstatobs}{M}
\newcommand{\forallhumanslongset}{\set{1,\dots,\numhumans}}

\newcommand{\forallhumans}{\forall \idA \in \set{1,\dots,\numhumans}}

\newcommand{\forallstatobsshortset}{\set{\numhumans+1,\dots,\numhumans+\numstatobs}}

\newcommand{\actionhum}{\tilde{\control}}
\newcommand{\actionhumat}[1]{{\at{\actionhum}{#1}}}

\newcommand{\idA}{j}
\newcommand{\idB}{l}
\newcommand{\idStat}{\tilde{l}}

\newcommand{\orcarlxsolnset}{\mathcal{O}}

\newcommand{\horiz}{T}

\newcommand{\stagecostsymb}{l}
\newcommand{\stagecost}[1]{\stagecostsymb(#1)}

\newcommand{\termpenalsymb}{\stagecostsymb_{\horiz}}
\newcommand{\termpenal}[1]{\termpenalsymb(#1)}

\newcommand{\forallactidcs}{\forall t\in\set{0,\dots,\horiz-1}}

\begin{acronym}[SICNav]
    \acrodef{EKF}{Extended Kalman Filter}
    \acrodef{KF}{Kalman Filter}
    \acrodefplural{KFs}{Kalman Filters}
    \acro{MPC}{Model Predictive Control}
    \acro{CVAE}{Conditional Variational Autoencoder}
    \acro{DWA}{Dynamic Window Approach}
    \acro{EB}{Elastic Band}
    \acro{FRP}{Freezing Robot Problem}
    \acro{GP}{Gaussian Process}
    \acrodefplural{GP}{Gaussian Processes}
    \acro{IGP}{Interacting Gaussian processes}
    \acro{IL}{Imitation Learning}
    \acro{IRL}{Inverse Reinforcement Learning}
    \acro{LSTM}{Long-Short Term Memory}
    \acrodefplural{LSTM}{Long-Short Term Memory networks}
    \acro{MAP}{Maximum a Posteriori}
    \acro{MLP}{Multi-Layer Perceptron}
    \acro{MPC}{Model Predictive Control}
    \acro{ORCA}{Online Reciprocal Collision Avoidance}
    \acro{PCLRHC}{Partially Closed Loop Receding Horizon Control}
    \acro{RHC}{Receding Horizon Control}
    \acro{RL}{Reinforcement Learning}
    \acro{ROS}{Robot Operating System}
    \acro{SARL}{Socially Attentive Reinforcement Learning}
    \acro{RGL}{Relational Graph Learning}
    \acro{SOGM}{Spatiotemporal Occupancy Grid Map}
    \acro{TEB}{Timed Elastic Band}
    \acro{VAE}{Variational Autoencoder}
    \acro{CVMM}{Constant Velocity Motion Model}
    \acro{MATS}{Mixtures of Affine Time-varying Systems}
    \acro{AV}{Autonomous vehicle}
    \acrodefplural{AVs}{Autonomous vehicles}

    \acrodef{LP}{Linear Program}
    \acrodef{QCQP}{Quadratically Constrained Quadratic Program}
    \acrodef{MPCC}{Mathematical Program with Complementarity Constraints}
    \acrodef{KKT}{Karush-Kuhn-Tucker}
    \acrodef{LICQ}{Linear Independence Constraint Qualification}
    \acrodef{CRCQ}{Constant Rank Constraint Qualification}
    \acrodef{SCQ}{Slater's Constraint Qualification}

    \acrodef{OGM}{Occupancy Grid Map}
    \acrodef{NF}{Navigation Function}
    \acrodef{RVO}{Reciprocal Velocity Obstacle}
    \acrodefplural{RVO}{Reciprocal Velocity Obstacles}
    \acrodef{SICNav}{Safe and Interactive Crowd Navigation}

    \acrodef{MPC}{Model Predictive Control}
    \acrodef{CAMPC}{Collision Avoiding Model Predictive Control}
    \acrodef{RHC}{Receding Horizon Control}
    \acrodef{CLF}{control-Lyapunov Function}
    \acrodef{DWA}{Dynamic Window Approach}
    \acrodef{SFM}{Social Force Model}
    \acrodef{ESFM}{Extended Social Force Model}

    \acrodef{ORCA}{Optimal Reciprocal Collision Avoidance}
    \acrodef{VO}{Velocity Obstacle}
    \acrodef{CA}{Collision Avoiding}

    \acrodef{ADE}{Average Displacement Error}
    \acrodef{FDE}{Final Displacement Error}
    \acrodef{KDE}{Kernel Density Estimate}
    \acrodefplural{KDE}{Kernel Density Estimates}
    \acrodef{NLL}{Negative Log Likelihood}

    \acrodef{RMSE}{Root Mean Squared Error}

    \acro{SLAM}{Simultaneous Localization and Mapping}
    \acro{VNC}{Virtual Network Computing}
    \acro{IMU}{Intertial Measurement Unit}
    \acro{CAD}{Computer Aided Design}
\end{acronym}

\title{Deploying SICNav in the Field: Safe and Interactive Crowd Navigation using MPC and Bilevel Optimization}

\author{Sepehr Samavi$^{1,2}$, Garvish Bhutani$^{2}$, Florian Shkurti$^{2}$, Angela P. Schoellig$^{1,2}$%
\thanks{$^{1}$Learning Systems and Robotics Lab at the Technical University of Munich and Munich Institute for Robotics and Machine Intelligence, Munich, Germany.%
$^{2}$University of Toronto Robotics Institute and the Vector Institute for Artificial Intelligence, Toronto, Canada. %
Emails: {\tt\scriptsize s.samavi@utoronto.ca, garvish.bhutani@mail.utoronto.ca, florian@cs.toronto.edu, angela.schoellig@tum.de}.}%
}
\usepackage{fancyhdr}
\fancypagestyle{withfooter}{
  
  \fancyhead[L]{}
  \fancyhead[R]{}
  \fancyfoot[C]{\footnotesize Presented at the 2025 IEEE ICRA Workshop on Field Robotics}
}
\begin{document}
\maketitle
\begin{abstract}
  Safe and efficient navigation in crowded environments remains a critical challenge for robots that provide a variety of service tasks such as food delivery or autonomous wheelchair mobility. Classical robot crowd navigation methods decouple human motion prediction from robot motion planning, which neglects the closed-loop interactions between humans and robots. This lack of a model for human reactions to the robot plan (e.g. moving out of the way) can cause the robot to get stuck.
  Our proposed Safe and Interactive Crowd Navigation (SICNav) method is a bilevel Model
  Predictive Control (MPC) framework that combines prediction
  and planning into one optimization problem, explicitly modeling interactions among agents. 
  In this paper, we present a systems overview of the crowd navigation platform we use
  to deploy SICNav in previously unseen indoor and outdoor environments. We
  provide a preliminary analysis of the system’s operation over the course of nearly 7 km of autonomous navigation over two hours in both indoor and outdoor environments.
\end{abstract}
\thispagestyle{withfooter}
\pagestyle{withfooter}

\section{Introduction}
Navigating among human crowds is a critical step towards deploying robots that offer a variety of services, such as autonomous food delivery or wheelchair operation. However, this is a challenging task for robots due to the uncertainty of pedestrian intentions and motions. Conventional approaches use decoupled frameworks (e.g  \cite{mayne2000constMPC,DuToit2012}) where human motion prediction (e.g. \cite{saltzmann2021trajpp,mangalam2021ynet,yue2022nspsfm}) is independent of planning, leading to the robot freezing and potential collisions. Recent interactive strategies (e.g. \cite{sun_move_2021, chen_relational_2020,Everett2021}) leverage closed-loop frameworks, but must rely on post hoc safety filters to guarantee non-collision, potentially resulting in oscillations or chatter.
\begin{figure}[t]
	\centering
	\footnotesize
		\centering
		\includegraphics[width=\linewidth]{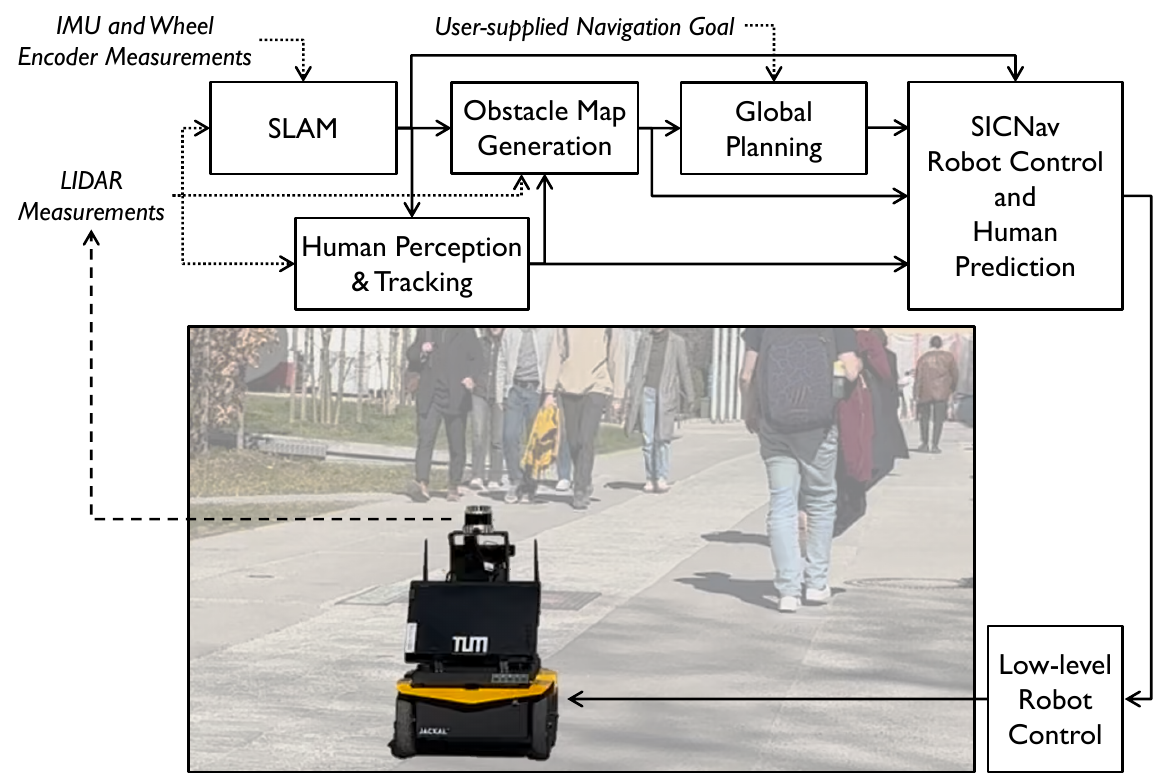}
    \caption{Photo of the robot operating in a crowd environment superimposed with a block-diagram of the components in the autonomy stack. Video of pictured robot operating autonomously: \href{https://tiny.cc/sicnav_field_ws}{\texttt{tiny.cc/sicnav\_field\_ws}}}\label{fig:block_diag}
\end{figure}

This paper presents the initial field deployment of SICNav \cite{samavi2024sicnav}, a bilevel nonlinear MPC framework, which integrates ORCA-modeled human predictions \cite{vandenBerg2011orca} directly into the robot's local planner. SICNav ensures collision-free trajectories while explicitly modeling interactions by jointly optimizing robot plans and human trajectory predictions. Unlike the aforementioned interactive methods, SICNav generates joint robot plans and human predictions that are collision-free by construction, ensuring safety while maintaining interactive behavior.

In this paper, we present the systems overview of the crowd navigation platform we use to deploy SICNav on the Clearpath Jackal robot pictured in Fig.~\ref{fig:block_diag}. The target task is to place a robot in a priori unknown planar environment with humans and static obstacles, then provide a goal location to the robot and have it navigate to the goal while avoiding collisions with humans and static obstacles.

We build upon existing robot autonomy components to enable our robot to navigate in previously unseen indoor and outdoor environments.  As illustrated in Fig.~\ref{fig:block_diag}, our system integrates 2D \ac{SLAM} based on Google Cartographer \cite{hess2016cartograper}, human perception and tracking based on YOLOv9 to generate 2D detections \cite{wang2024yolov9} with aUToTrack for human tracking \cite{burnett2019autotrack}, and obstacle map generation and global path planning using the \ac{ROS} 1 \texttt{move\_base} framework. Our method, SICNav is used instead of the default local planner in \texttt{move\_base} to jointly generate robot trajectories and human predictions that are collision-free and interactive with humans. We use the SICNav solutions in receding horizon fashion by sending velocity commands to the robot base, which are tracked by the robot's onboard low-level controllers.

Our contributions are threefold. First, we integrate the aforementioned components of the robot autonomy stack to deploy SICNav in previously unseen indoor and outdoor environments, with noisy state estimation for robot localization, as well as detection and tracking. Second, we propose algorithmic improvements to SICNav and the detection and tracking components to improve the performance of the system. Finally, we provide a preliminary analysis of the system's operation in the field and present our plan to thoroughly evaluate the algorithm's performance in the final version of this paper.

\section{Methodology}
\begin{figure}[t]
	\centering
	\footnotesize
		\centering
		\includegraphics[width=\linewidth]{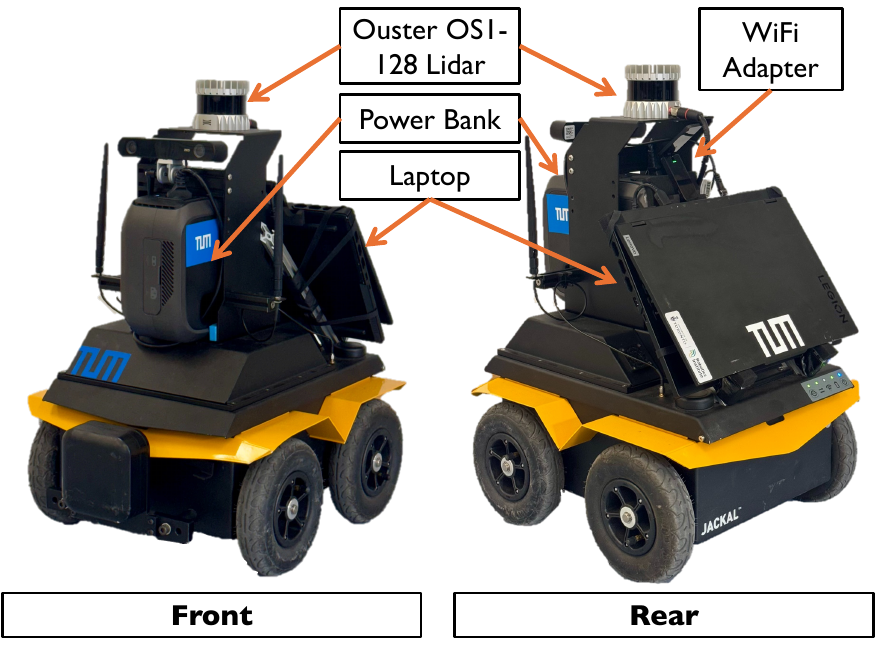}
    \caption{Photos of the front (left) and rear (right) view of the Clearpath Jackal robot platform with labels for components used in this paper.}\label{fig:robot_diag}
\end{figure}
\subsection{The Robot}
The robot used in our experiments, illustrated in Fig.~\ref{fig:robot_diag}, is a Clearpath Jackal four-wheeled slip steer robot weighing $20kg$ and measuring $46cm$ W $\times 60cm$ L $\times 50cm$ H. We use an Ouster OS1-128 Lidar sensor, its built-in IMU, and the built-in wheel encoders of the robot as the only sensors in this project. The robot is equipped with an onboard computer running Ubuntu 20.04 and ROS 1 Noetic. We additionally mount a laptop with 24-core Intel Core i9-14900HX CPUs and an Nvidia RTX 4080 GPU to the robot.

In order to enable communication with the robot to monitor visualization and issue commands (e.g. specifying goal location, recording data), we use a TP-Link T4U WiFi USB-adapter to broadcast a WiFi network from the laptop. The operator connects to this network using a hand-held laptop or a tablet and uses \ac{VNC} screen sharing to interact with ROS and the RVIZ visualization wirelessly.

\subsubsection{Sensor Data Preprocessing} We pre-process the lidar data by employing Patchwork++ \cite{lee2022patchwork} to segment ground points from the lidar pointcloud used for mapping and localization as well as obstacle detection. The algorithm proposes a ground likelihood estimation method to segment the ground from the point cloud, tracking the ground plane over time to improve the segmentation. Removing the ground points from the point cloud simplifies our mapping and obstacle detection tasks by reducing the number of points to process.

\subsection{Cartographer SLAM} \label{sec:slam}
We employ the 2D version of Google Cartographer \cite{hess2016cartograper} for \ac{SLAM} in our autonomy stack. Our setup relies on lidar point clouds, the lidar's \ac{IMU} measurements, and wheel odometry supplied by the robot base. The Cartographer algorithm first collapses the point cloud into a 2D scan in birds-eye-view, using the \ac{IMU} to estimate the direction of gravity and projecting down. The method then uses scan matching to align incoming lidar scans with generated submaps consisting of a few consecutive scans. To improve scan matching performance, the algorithm incorporates linear and angular velocities from the most recent two odometry readings to help in extrapolating the pose when aligning the incoming lidar scans with the map. To align the submaps into one consistent map, the system constructs a pose graph of all scans and submaps and periodically conducts loop closure optimizations.

The pose estimation is conducted in the frame of the lidar \ac{IMU}. We eliminate the need for a lidar-to-\ac{IMU} calibration by using the lidar's \ac{IMU}. To transform the odometry data into the \ac{IMU} frame, we use a static transformation obtained from the \ac{CAD} drawings of the robot base. We further filter the pose estimates published by Cartographer using a \ac{EKF} to estimate the robot's velocity and acceleration which are used in the SICNav optimization problem, described in Sec.~\ref{sec:sysstatedyn}.
\subsection{Human Detection and Tracking}
\begin{figure}
  \begin{subfigure}{\linewidth}
    \centering
    \includegraphics[width=\linewidth]{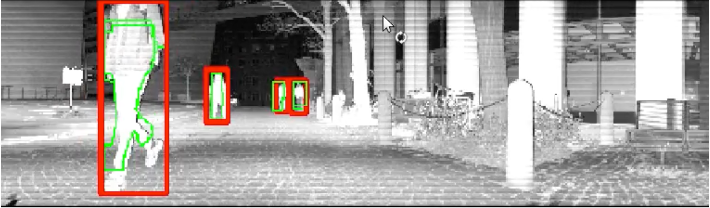}
    \caption{Signal (Intensity) Image}
    \label{fig:yolodet_img}
  \end{subfigure}
  \begin{subfigure}{\linewidth}
    \centering
    \includegraphics[width=\linewidth]{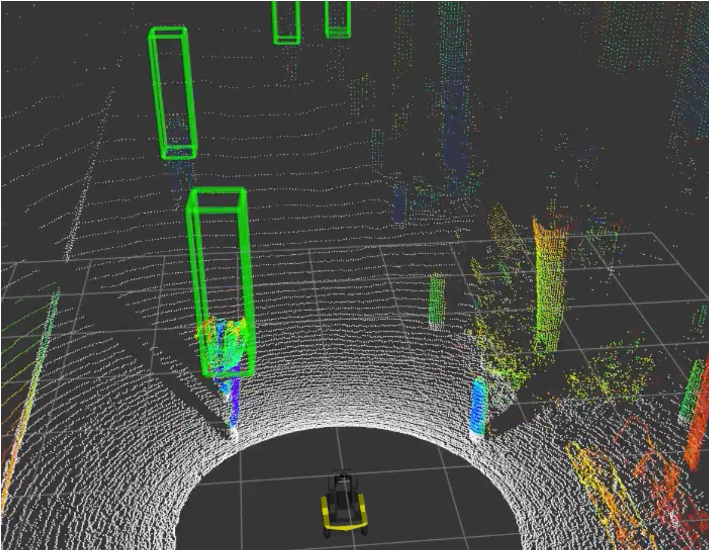}
    \caption{Point cCloud}
    \label{fig:yolodet_pts}
  \end{subfigure}
  \caption{Lidar signal image (a) and point cloud (b) generated by the Ouster OS1-128 lidar sensor. In (a), we overlay Yolov9 detection bounding boxes (red) and segmentation contours (green). In (b) we illustrate the 3D positions associated with the detections with bounding boxes (green) with the height augmented for easier visualization.} \label{fig:yolodet}
\end{figure}
We adapt the aUToTrack human detection and tracking methodology \cite{burnett2019autotrack}. The original method obtains 2D bounding boxes on a camera image, then uses a lidar-to-camera calibration to associate the lidar points associated with each bounding box. The method conducts Euclidean clustering to the points associated with each bounding box to obtain a 3D centroid estimate for each bounding box. Since the bounding boxes and lidar points are in different frames, this method is sensitive to poor calibration between the lidar and the camera.

We alter the aUToTrack method by using the signal (measuring intensity), reflectivity, and range (distance) images generated by the Ouster lidar. In these images, each pixel is associated with a point in the point cloud, eliminating the need for a camera altogether. In order to obtain detections, we use the segmentation variant of Yolov9 \cite{wang2024yolov9} on the signal and reflectivity images to obtain a contour corresponding to each detection. Fig.~\ref{fig:yolodet_img} illustrates the detections on the intensity image. The contours may contain pixels from the ground or other objects. Thus, in order to reject points that do not belong to the detected object we analyze the range values of the pixels in the detected contour and reject any pixels with range below the $5^{th}$ percentile and above the $65^{th}$ percentile of the pixels. From the remaining pixels, we calculate the Euclidean mean of the 3D positions of the detection. Fig.~\ref{fig:yolodet_pts} illustrates the positions of the detections in the pointcloud.

Once we obtain a set of detections, we use the aUToTrack method for data association to maintain consistent identities for each detected human through successive frames. We use \acp{KF} for with a constant acceleration motion model for each tracked human.
Particularly for data association, we use the distance between the predicted state of the track and the 3D position of the detection and follow a greedy data association technique. In order to compensate for tracking delay, we use the constant acceleration model to project the \ac{KF} estimate forward in time, which we then use in the SICNav optimization problem.

\subsection{Obstacle Map Generation and Global Path Planning} \label{sec:costmaps}
We set up the ROS \texttt{move\_base} framework with a global and local \textit{Costmap2D}. The framework obtains the map from the Cartographer node then uses lidar measurements to generate a 2D occupancy grid. We filter out point-cloud data associated with moving humans, using their tracks to exclude them from the global map. This process allows SICNav to address the humans within its optimization framework. The global occupancy map is used for global planning with the hybrid A* algorithm, which finds a path for the robot to take to its user-supplied goal by blending discrete search with continuous state evaluations. For the local cost map, we use the ROS 1 \textit{costmap\_converter} package \cite{rosmann2015costmapconverter} to cluster points in the occupancy grid into polygons, which are then translated into line segment representations. We use these segments in SICNav's static obstacle collision avoidance constraints.

\subsection{SICNav}
\subsubsection{Problem Formulation}
The environment consists of a robot, human agents, indexed by $\idA \in \forallhumanslongset$, and static obstacles in the form of line segments, indexed by $\idStat \in \forallstatobsshortset$.
\label{sec:sysstatedyn}
The state of the system is in continuous space and contains the state of the ego robot, and the states of all $\numhumans$ humans.
The state is denoted as
$
  \stateat{t} = (\robstateat{t},    \id{1}{\humstateat{t}}, \dots,  \id{\numhumans}{\humstateat{t}})  \in \sspace,
$
where the state of the robot,
$\robstateat{t} \tuple{\xpos_t, \ypos_t, \head_t, \lvel_t, \avel_t, \dot{\lvel}_t, \dot{\avel}_t} \in \Reals^{7}$,
consists of its 2D position, heading, and longitudinal velocity, angular velocity, longitudinal acceleration, and angular acceleration. %
The human state,
$\id{\idA}{\humstateat{t}} \in \Reals^{4}$,
consists of 2D position, and 2D velocity $\forallhumans$.

We separate the dynamics of the system into separate functions for the robot and the humans. The robot dynamics,
${\robstateat{\kpone}} = \robdyn({\robstateat{t}}, \actionat{t}),$
are modeled as a kinematic unicycle model, where the control input for the system, $\at{\action}{t} \in \cspace$, is a vector of the linear and angular velocity of the robot for time step $t$.
The dynamics of each human $\forallhumans$,
$\id{\idA}{\humstateat{\kpone}} = \humdyn(\id{\idA}{\humstateat{t}}, \id{\idA}{\actionhumat{t}}),$
are modeled as kinematic integrators, where $\id{\idA}{\actionhumat{t}}$ denote actions optimal with respect to \ac{ORCA}.

\subsubsection{ORCA for Predicting Human Motion}
The \ac{ORCA} optimization problem for agent $\idA$ at some time $t$ is a \ac{QCQP} that solves for a predicted human velocity that minimizes the distance to a intended velocity,
\begin{equation} \label{eq:relaxed_orca_argmin}
    \id{\idA}{\orcarlxsolnset}(\at{\state}{t}) := \argmin_{{\vel}} \\ \left\{
    \begin{alignedat}{0}
    \centremathcell{\twonorm{{\vel} - {{\vel_{\pref}}}(\stateat{t})}^2 \text{}:\text{}}\\
    \centremathcell{\text{collision-avoidance}}
    \end{alignedat}\right\},
\end{equation}
where the optimization variable $\vel \in \Reals^2$ is the velocity of agent $j$ for one time step. The cost is based on a velocity vector from the agent's current position towards the agent's intended goal position, ${\id{\idA}{\mathbf{g}}}$, which we estimate by projecting the agent's latest observed velocity forward with a fixed time horizon. The non-collision constraints are derived using velocity obstacle approach (see Sec. 4 of \cite{vandenBerg2011orca}) and are composed of a linear constraint for agent $j$ and each other agent $l \neq j$, as well as each static obstacle. These constraints are linear with respect to $\vel$ and parameterized by the current state of the system, $\stateat{t}$. The problem also contains a convex quadratic maximum velocity constraint.

The bilevel optimization problem uses the optimum of \eqref{eq:relaxed_orca_argmin}, $\id{\idA}{\actionhumat{t}}$, to find the refined predicted position of the agent at the next time step, $\id{\idA}{\at{\humstate}{\kpone}}$.
This way, if the intended velocity for the agent is feasible with respect to constraints in \eqref{eq:relaxed_orca_argmin}, then the optimal velocity found by solving the problem is the one that moves the agent to the estimated intended position at the next time step.
\begin{figure*}[th]
	\centering
	\footnotesize
  \begin{subfigure}[]{0.49\textwidth}
    \centering
		\includegraphics[width=\linewidth]{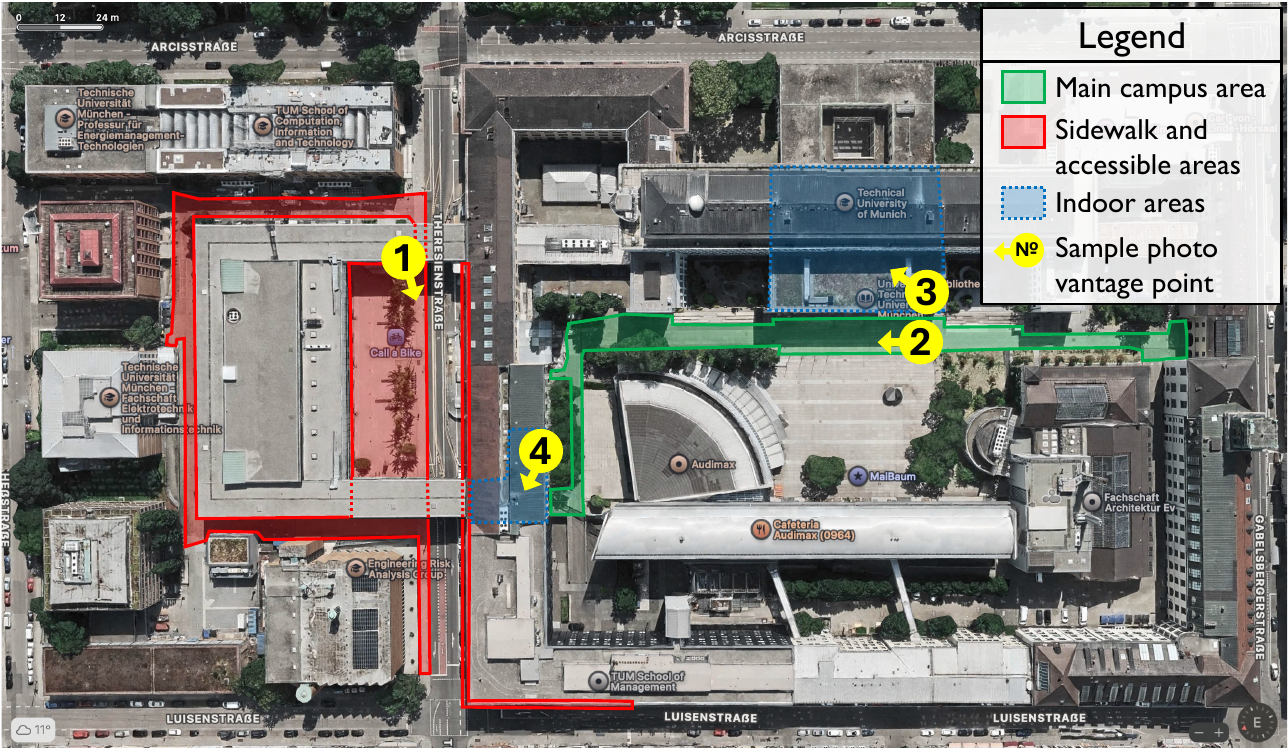}
    \caption{Map of robot operation areas} \label{fig:areas_map}
  \end{subfigure}
  \begin{subfigure}[]{0.49\textwidth}
    \centering
		\includegraphics[width=0.95\linewidth]{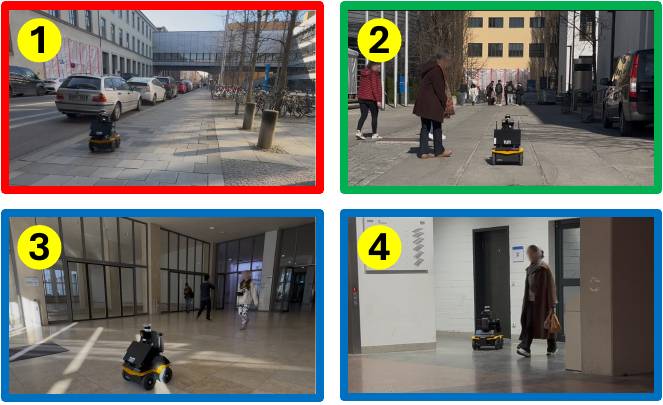}
    \caption{Photos of robot operation areas}\label{fig:areas_photos}
  \end{subfigure}
  \caption{(a) Robot operation areas highlighted on the Apple Maps satellite image of the Technical University of Munich campus in Maxvorstadt, Munich, Germany. (b) Photos of the robot operating in each area.}\label{fig:areas}
\end{figure*}
\subsubsection{SICNav Bilevel Optimization Problem} \label{sec:sicnav}
The bilevel optimization problem is formulated as follows,
\begin{subequations} \label{eq:combmpc0_prob}
	\begin{alignat}{2}
		\centremathcell{\minimize_{\substack{\stateat{0:\horiz},\actionat{0:\horiz-1},\\\id{0:\numhumans}{\at{\actionhum}{0:\horiz-1}}}}} 	& \centremathcell{\sum_{t=0}^{\horiz-1} \stagecost{\stateat{t},\actionat{t}} + \termpenal{\stateat{\horiz}}} 		& \centremathcell{\quad\quad\quad\quad\quad} \label{eq:combmpc0_min}			\\
		\centremathcell{\subjectto} 																			  							& \centremathcell{\stateat{0}=\stateinit} 																	 		& \centremathcell{\quad\quad\quad\quad\quad} \label{eq:combmpc0_initcondconst}	\\
		\centremathcell{}																										 			& \centremathcell{\rob{\stateat{\kpone}}= \robdyn(\rob{\stateat{t}},\controlat{t})} 							     		& \centremathcell{\quad\quad\quad\quad\quad} \label{eq:combmpc0_rob_dynconst}	\\ 	
		\centremathcell{}																										 			& \centremathcell{\id{\idRob}{\action_{min}} \leq {\at{\action}{t}} \leq \id{\idRob}{\action_{\max}}} 	 		& \centremathcell{\quad\quad\quad\quad\quad} \label{eq:combmpc0_actionconst0}	\\ 	
		\centremathcell{}																										 			& \centremathcell{\Delta\id{\idRob}{\action_{min}} \leq {\at{\action}{t} - \at{\action}{\kmone}} \leq \Delta\id{\idRob}{\action_{\max}}} 	 		& \centremathcell{\quad\quad\quad\quad\quad} \label{eq:combmpc0_actionconst}	\\ 	
		\centremathcell{}																										 			& \centremathcell{\trans{\at{\state}{t}} \P_\idB \at{\state}{t} \geq {\dist_{\idB}}^2}   	& \centremathcell{\quad\quad\quad\quad\quad} \label{eq:combmpc0_coll_const}	\\ 	
	  	\centremathcell{}																										 			& \centremathcell{\id{\idA}{\at{\actionhum}{t}} \in \id{\idA}{\orcarlxsolnset}(\at{\state}{t})}				 		& \centremathcell{\quad\quad\quad\quad\quad} \label{eq:combmpc0_llorca}		\\ 	
	  	\centremathcell{}																										 			& \centremathcell{\id{\idA}{\stateat{\kpone}} = \humdyn(\id{\idA}{\stateat{t}}, \id{\idA}{\at{\actionhum}{t}})}	 		& \centremathcell{\quad\quad\quad\quad\quad} \label{eq:combmpc0_hum_dynconst}		
	\end{alignat}
\end{subequations}
where the constraints for robot dynamics \eqref{eq:combmpc0_rob_dynconst}, robot velocity bounds, \eqref{eq:combmpc0_actionconst0} and robot acceleration bounds \eqref{eq:combmpc0_actionconst} are defined for each time step, $\forallactidcs$; the robot non-collision constraint \eqref{eq:combmpc0_coll_const} is defined for each human agent and each static obstacle in the environment; and the lower-level ORCA optimization problem constraint \eqref{eq:combmpc0_llorca} and human dynamics constraint \eqref{eq:combmpc0_hum_dynconst} are defined for each human, $\forallhumans$ and each time step.
To solve this bilevel problem, we reformulate to a single level by replacing each lower level problem with its \ac{KKT} optimality conditions.

We solve the resulting optimization problem using the Acados nonlinear programming solver \cite{verschueren2021acados}. We use the solution in receding horizon fashion to generate velocity commands for the robot base, which are tracked by the Jackal's onboard low-level controllers. In the \ac{MPC} problem, we use a discrete time step of $\deltat = 0.25s$, a prediction horizon of $\horiz = 2 s$, and we re-plan at $10Hz$. To ensure the problem is solved in real-time, we model the two humans with the lowest time-to-collision with the robot using \ac{ORCA}. In order to avoid interference from the global planner, we remove the lidar points associated with these humans from the point cloud such that these humans do not appear as obstacles in the costmaps described in Sec.~\ref{sec:costmaps}. This way, we treat any additional humans as static obstacles.

\section{Field Tests}
\begin{figure}[t]
	\centering
	\footnotesize
  \centering
  \includegraphics[width=\linewidth]{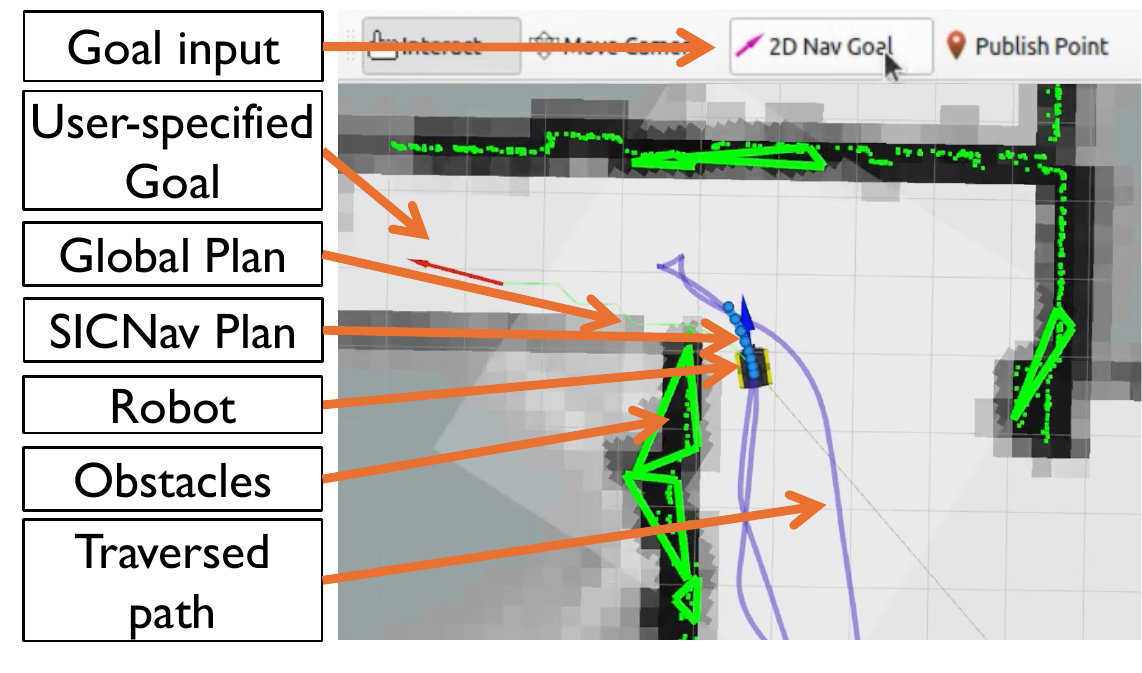}
  \caption{Top-down view of the Rviz visualization and goal input interface available to the user. The visualization shows a top-down view of the robot's pose in the map that has been constructed using \ac{SLAM}. In this example the robot is in the operation area labelled 4 in Fig.~\ref{fig:areas_photos}. The user specifies goal positions a goal to the robot and monitor robot plans.} \label{fig:command_interface}
\end{figure}

\begin{figure*}
  \begin{subfigure}{0.32\textwidth}
    \centering
    \includegraphics[width=\linewidth]{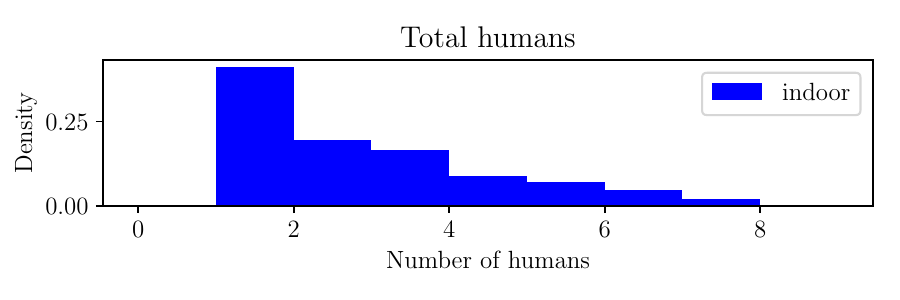}    \caption{Indoor areas}
    \label{fig:tothums_indoor}
  \end{subfigure}
  \begin{subfigure}{0.32\textwidth}
    \centering
    \includegraphics[width=\linewidth]{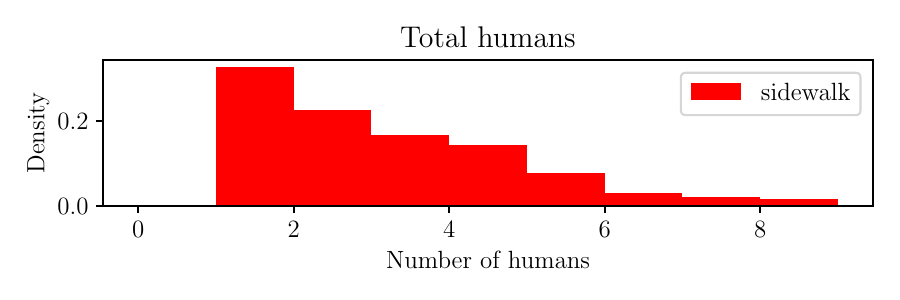}
    \caption{Sidewalk and Accessible Areas}
    \label{fig:tothums_sidewalk}
  \end{subfigure}
  \begin{subfigure}{0.32\textwidth}
    \centering
    \includegraphics[width=\linewidth]{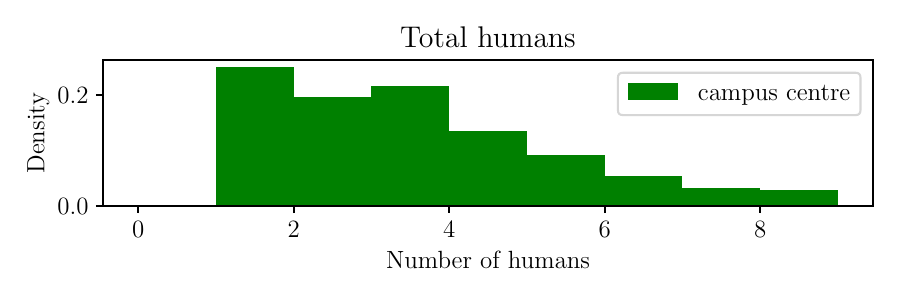}
    \caption{Main Campus Area}
    \label{fig:tothums_campus}
  \end{subfigure}
  \caption{Total number of humans detected in the scene in the different operation areas. We plot histograms for the (a) Indoor areas $\mu=2.450$, $\sigma=1.650$; (b) Sidewalk and accessible areas $\mu=2.716$, $\sigma=1.796$; and (c) Campus center areas $\mu=3.128$, $\sigma=1.986$. We observe that the main campus area had the largest crowds followed by the sidewalk and accessible areas and the indoor areas had the lowest crowds.} \label{fig:tothums}
\end{figure*}

\begin{figure*}
  \begin{subfigure}{0.32\textwidth}
    \centering
    \includegraphics[width=\linewidth]{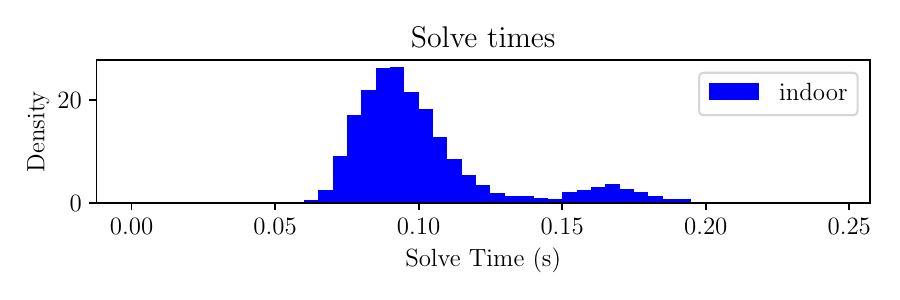}
    \caption{Indoor areas}
    \label{fig:solvetimes_indoor}
  \end{subfigure}
  \begin{subfigure}{0.32\textwidth}
    \centering
    \includegraphics[width=\linewidth]{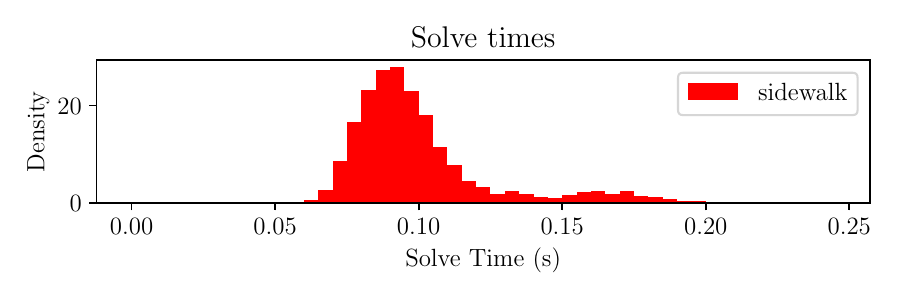}
    \caption{Sidewalk and Accessible Areas}
    \label{fig:solvetimes_sidewalk}
  \end{subfigure}
  \begin{subfigure}{0.32\textwidth}
    \centering
    \includegraphics[width=\linewidth]{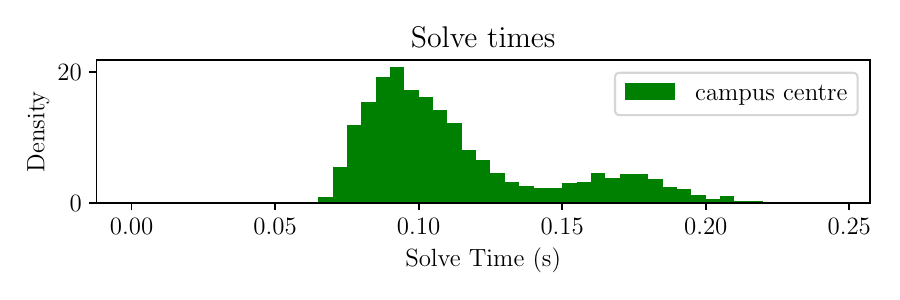}
    \caption{Main Campus Area}
    \label{fig:solvetimes_campus}
  \end{subfigure}
  \caption{SICNav solve times in the (a) Indoor areas $\mu=0.102$, $\sigma=0.027$; (b) Sidewalk and accessible areas $\mu=0.100$, $\sigma=0.026$; and (c) Campus center areas $\mu=0.114$, $\sigma=0.035$. All areas had acceptable solve times i.e. lower than discretization period of 0.25. We observe that the indoor and sidewalk areas had similar solve times while the main campus area had larger times.} \label{fig:solvetimes}
\end{figure*}
\subsection{Deployment Methodology}
We have begun deploying the robot in the Technical University of Munich campus in Maxvorstadt, Munich, Germany. The campus has a variety of environments, including narrow hallways, open courtyards, and outdoor pathways. The results presented in this paper are preliminary. In Sec.~\ref{sec:nextsteps} we will highlight the additional field results we plan to include in the final version of the paper.

We have identified three areas for robot operation, as shown in Fig.~\ref{fig:areas}. The \textit{main campus area} (green) is an outdoor pedestrian zone separated from city roads by university buildings. The surfaces in this area are covered variably by cobblestones, asphalt, and pavement. The environment contains a variety of obstacles such as light poles, benches, bicycle parking, and parked cars. The \textit{sidewalk and accessible areas} (red) are the sidewalks and pedestrian zones adjacent to university buildings and accessible directly adjacent to city roads. This area is similar to the main campus area but additionally contains moving cars on the road adjacent to the sidewalk. The \textit{indoor areas} (blue) are located within two university buildings. The surfaces of the indoor areas are stone flooring with some carpets. The environment contains narrow hallways, staircases, and open spaces with benches and other furniture.

To operate the robot once it is in one of the operation areas, we connect wirelessly to the robot laptop using a hand-held laptop or tablet. Then we view the map that has been collected by the robot (as described in Sec.~\ref{sec:slam}) in Rviz, as illustrated in Fig.~\ref{fig:command_interface}. To commence navigation, we specify a goal pose on the map using the 2D Nav Goal interface. Once the robot reaches the pose, it stops operating.

\subsection{Results}
\begin{table}[t]
  \centering
  \footnotesize
  \caption{Autonomous Robot Operation Statistics}
  \label{tab:compact_drive_statistics}
  \begin{tabular}{|l|ccc|c|}
  \hline
  & Campus & Sidewalk & Indoor & Total \\
  \hline
  Duration (h:m:s)& 0:57:26 & 0:37:04 & 0:17:39 & 1:51:41 \\
  Distance (km) & 3.75 & 2.02 & 0.96 & 6.73 \\
  \# Manual TOs & 20 & 27 & 2 & 49 \\
  TO Freq. (s$^{-1}$)& 0.0058 & 0.0121 & 0.0019 & 0.0073 \\
  \hline
  \end{tabular}
 \end{table}

Table~\ref{tab:compact_drive_statistics} summarizes the robot's operation statistics in the three areas. The robot has navigated autonomously for nearly one hour and 51 minutes, traversing 6.73 km across the three environments. A video of the robot navigating in the three areas can be found at \href{https://tiny.cc/sicnav_field_ws}{\texttt{tiny.cc/sicnav\_field\_ws}}.

To evaluate robot performance, we first focus on manual takeovers. The sidewalk environment had the highest number of takeovers, totaling 27, with a frequency of 0.0121 s\textsuperscript{-1}. We believe this increase is due to the current setup's inability to detect steps going down, causing the robot to sometimes attempt driving off the sidewalk and requiring manual intervention. In contrast, the campus and indoor environments had fewer takeovers, with frequencies of 0.0058 s\textsuperscript{-1} and 0.0019 s\textsuperscript{-1}, respectively.

We also evaluate the number of humans detected by the robot during operation. Fig.~\ref{fig:tothums} shows the total number of humans detected in the different operation areas. The main campus area had the largest crowds, followed by the sidewalk and accessible areas, while the indoor areas had the lowest crowds. The robot detected an average of 2.45 humans in the indoor areas, 2.72 humans in the sidewalk and accessible areas, and 3.13 humans in the main campus area.

Finally, we evaluate the solve-times of the SICNav optimization problem in the different areas, summarized in Fig.~\ref{fig:solvetimes}. We observe that the indoor and sidewalk areas have similar solve times, while the main campus area had larger times. We believe this is as a result of the higher number of humans encountered in the problem. All areas had acceptable solve times, lower than the discretization period described of 0.25 seconds used in the SICNav \ac{MPC} problem (see Sec~\ref{sec:sicnav}).

\subsection{Current Limitations and Next Steps} \label{sec:nextsteps}
\begin{figure*}
  \begin{subfigure}{0.49\textwidth}
    \centering
    \includegraphics[width=\linewidth]{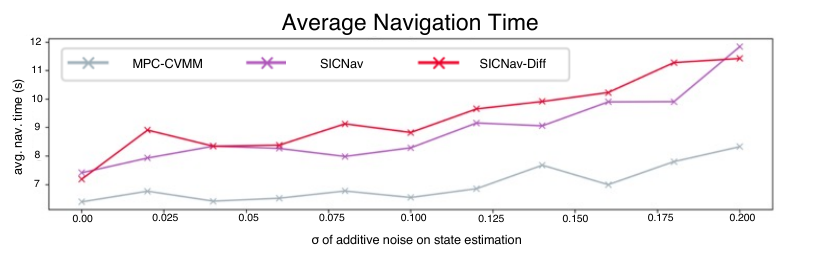}
    \caption{Navigation Time}
    \label{fig:sens_nav_time}
  \end{subfigure}
  \begin{subfigure}{0.49\textwidth}
    \centering
    \includegraphics[width=\linewidth]{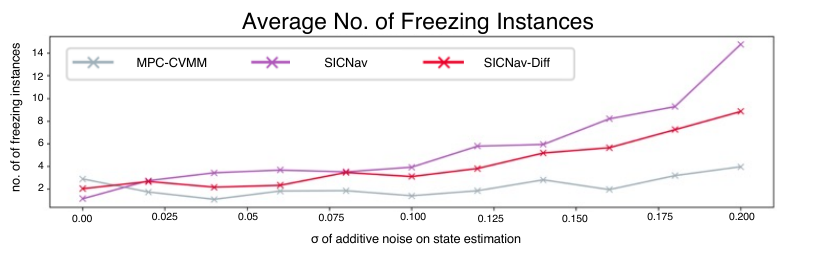}
    \caption{Robot Freezing Instances}
    \label{fig:sens_freez}
  \end{subfigure}
  \begin{subfigure}{0.49\textwidth}
    \centering
    \includegraphics[width=\linewidth]{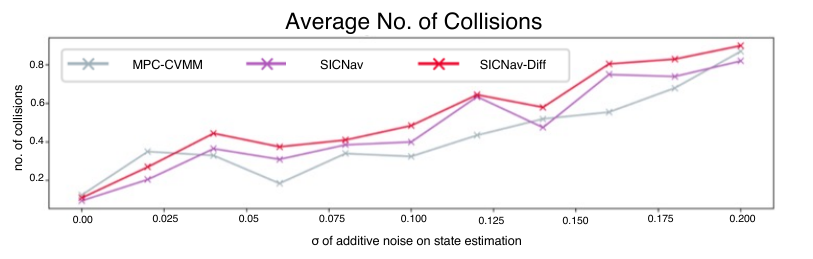}
    \caption{Robot-Human Collisions}
    \label{fig:sens_colls}
  \end{subfigure}
  \begin{subfigure}{0.49\textwidth}
    \centering
    \includegraphics[width=\linewidth]{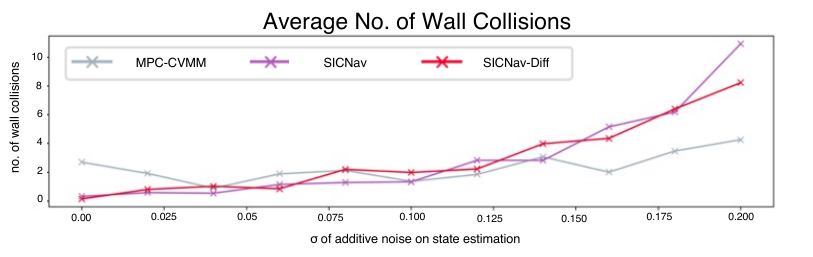}
    \caption{Robot-Wall Collisions}
    \label{fig:sens_wall_colls}
  \end{subfigure}
  \caption{Robot crowd navigation performance using different control methods in the face of increasing state estimation noise in simulation. We plot average (a) navigation time, (b) robot freezing instances, (c) number of robot-human collisions, and (d) robot wall collisions over 200 scenarios. When the standard deviation of additive noise increases beyond 0.03 the differences between the methods decreases.} \label{fig:sensitivity}
\end{figure*}
\subsubsection{SICNav Performance Comparison with Other Methods}
We have so far 
demonstrated the operation of SICNav in the field. However, in the final version of this paper, we aim to compare the performance of SICNav with other local planning methods in terms of navigation efficiency metrics such as time to goal, distance traveled, and number of manual takeovers.
The version of \ac{SICNav} that we have deployed uses the velocity estimate of the human to predict the intended direction of the human, then uses the \ac{ORCA} method to predict the humans motion towards this intent while jointly solving for the robot plan. The first method we intend to compare with is \ac{SICNav}-Diffusion \cite{samavi2025sicnavdiff} which uses a data-driven diffusion model to predict human intentions and incorporates these intention predictions into the optimization problem. We also intend to compare the method with a non-interactive MPC problem that uses a \ac{CVMM} to predict human trajectories without modelling interactions during robot planning.

In previous in-lab real-robot experiments \cite{samavi2024sicnav,samavi2025sicnavdiff} we have shown that SICNav outperforms the non-interactive MPC problem that uses a \ac{CVMM} to predict human trajectories in terms of metrics measuring robot navigation efficiency and safety. While in the in-lab experiments localization and perception components of the autonomy stack were provided by a VICON motion tracking system, in the field deployment, the robot operates with noisy localization and perception. One question that we aim to answer in the final paper is whether the differences in performance between SICNav and the other methods remain in the face of noisy localization and perception.

We have conducted a preliminary analysis of the relative performance of the methods under noisy localization and perception in simulation. We simulate a 1.75m wide corridor with $\numhumans = 3$ humans. We generate 200 scenarios with different initial and final positions on opposite sides of the corridor for each human. We evaluate three methods: SICNav, SICNav-Diffusion \cite{samavi2025sicnavdiff}, and the non-interactive MPC problem with \ac{CVMM} predictions. To evaluate performance we use average navigation time for the robot to reach its goal over the 200 scenarios, number of instances of robot freezing (where the robot stops moving), number of collisions with humans, and number of collisions with walls.
We repeat the 200 scenarios with varying levels of noise added to the robot position, robot velocity, human positions, human velocities, and static obstacle positions. We sample the noise from a zero-mean Gaussian distribution with standard deviation $\sigma \in \{0, 0.2\}$. We then evaluate the performance of the three methods under the noisy conditions.

The results are shown in Fig.~\ref{fig:sensitivity}. In Figs.~\ref{fig:sens_nav_time} and \ref{fig:sens_colls} we observe that as the standard deviation of the noise increases beyond 0.03, the relative performance of the methods begins changing, indicating that the noise in state estimation does not affect all the evaluated methods in the same way. In the final version of the paper, we aim to conduct a more thorough analysis of the performance of the methods under noisy conditions by varying the noise levels for each component of the autonomy stack separately and also incorporating delays in the perception and localization components of the autonomy stack.

\subsubsection{Additional Autonomy Components}
First, the robot's inability to detect steps going down other non-drivable areas (e.g. grass, traffic lanes) is a significant limitation that we aim to address by incorporating semantic segmentation of the lidar point cloud \cite{lai2023spherical} into the robot autonomy stack. This way we can detect non-drivable areas in the point cloud and include them as obstacles in the costmaps described in section~\ref{sec:costmaps}.

Second, %
In the current setup, if the user intends for the robot to drive to a location that is not currently in the robot's collected map, the user is required to manually specify an intermediate goal, then as the robot makes progress, update the goal once the final destination has appeared in the map. We aim to explore two options to address this limitation. First, we intend to explore collecting a map of the operation area of the robot prior to deployment, then accurately localize inside the map using the existing lidar \ac{SLAM} method that we currently use for \ac{SLAM}. If this approach does not scale to the large operating areas of the robot we then aim to explore incorporating a high-level task localization and planning component in addition to the current \ac{SLAM} and global planning components described in Sec.~\ref{sec:slam} and Sec.~\ref{sec:costmaps}. In the high-level localization module, we intend to use GPS for rough localization on a pre-collected map (e.g. Google Maps) and then use the lidar \ac{SLAM} method to localize the robot more accurately locally in the map. The high-level planning module will then use the robot's current pose in the map to generate waypoints to the goal location.

\subsubsection{Multi-city Testing}
So far, we have only tested the robot in the Technical University of Munich campus in Maxvorstadt, Munich, Germany. In the final version of this paper, we aim to test the robot in additional environments such as the city center of Munich, Germany, and the Technical University of Munich campus in Garching, Germany. We also have a similar robot base in the University of Toronto, Toronto, Canada. The second robot uses a lower-end lidar sensor (Ouster OS1-64) and lower-end on-board computer. We aim to test the robot in the University of Toronto campus and the city center of Toronto, Canada to evaluate the generalizability of the robot's performance in different environments and with different hardware configurations.

\section{Conclusion}
In this paper we presented the deployment of SICNav, our interactive crowd navigation framework based on bilevel optimization, on a Clearpath Jackal robot in and around the Technical University of Munich campus in Maxvorstadt, Munich, Germany. The robot has autonomously navigated for nearly two hours across three different environments, traversing 6.73 km over nearly two hours of autonomous operation. We demonstrated the ability of our robot to navigate in the face of noisy localization and perception in the field, and we presented a preliminary analysis of the performance of the robot under noisy conditions in simulation. In the final version of this paper, we aim to conduct a more thorough analysis of the performance of the robot under noisy conditions and compare the performance of SICNav with other local planning methods in terms of navigation efficiency metrics such as time to goal, distance traveled, and number of manual takeovers.

\bibliographystyle{IEEEtran}
\bibliography{IEEEabrv, references}

\begin{thebibliography}{10}
\providecommand{\url}[1]{#1}
\csname url@rmstyle\endcsname
\providecommand{\newblock}{\relax}
\providecommand{\bibinfo}[2]{#2}
\providecommand\BIBentrySTDinterwordspacing{\spaceskip=0pt\relax}
\providecommand\BIBentryALTinterwordstretchfactor{4}
\providecommand\BIBentryALTinterwordspacing{\spaceskip=\fontdimen2\font plus
\BIBentryALTinterwordstretchfactor\fontdimen3\font minus \fontdimen4\font\relax}
\providecommand\BIBforeignlanguage[2]{{%
\expandafter\ifx\csname l@#1\endcsname\relax
\typeout{** WARNING: IEEEtran.bst: No hyphenation pattern has been}%
\typeout{** loaded for the language `#1'. Using the pattern for}%
\typeout{** the default language instead.}%
\else
\language=\csname l@#1\endcsname
\fi
#2}}

\bibitem{mayne2000constMPC}
\BIBentryALTinterwordspacing
D.~Mayne, J.~Rawlings, C.~Rao, and P.~Scokaert, ``\BIBforeignlanguage{en}{Constrained model predictive control: {Stability} and optimality},'' \emph{\BIBforeignlanguage{en}{Automatica}}, vol.~36, no.~6, pp. 789--814, June 2000. [Online]. Available: \url{https://linkinghub.elsevier.com/retrieve/pii/S0005109899002149}
\BIBentrySTDinterwordspacing

\bibitem{DuToit2012}
\BIBentryALTinterwordspacing
N.~E. Du~Toit and J.~W. Burdick, ``{Robot {Motion} {Planning} in {Dynamic}, {Uncertain} {Environments}},'' \emph{IEEE Transactions on Robotics}, vol.~28, no.~1, pp. 101--115, Feb. 2012. [Online]. Available: \url{doi.org/10.1109/TRO.2011.2166435}
\BIBentrySTDinterwordspacing

\bibitem{saltzmann2021trajpp}
\BIBentryALTinterwordspacing
T.~Salzmann, B.~Ivanovic, P.~Chakravarty, and M.~Pavone, ``Trajectron++: {Dynamically}-{Feasible} {Trajectory} {Forecasting} {With} {Heterogeneous} {Data},'' in \emph{2020 {European} {Conference} on {Computer} {Vision} ({ECCV})}, 2020. [Online]. Available: \url{http://arxiv.org/abs/2001.03093}
\BIBentrySTDinterwordspacing

\bibitem{mangalam2021ynet}
\BIBentryALTinterwordspacing
K.~Mangalam, Y.~An, H.~Girase, and J.~Malik, ``\BIBforeignlanguage{en}{From {Goals}, {Waypoints} \& {Paths} {To} {Long} {Term} {Human} {Trajectory} {Forecasting}},'' in \emph{\BIBforeignlanguage{en}{2021 {IEEE}/{CVF} {International} {Conference} on {Computer} {Vision} ({ICCV})}}.\hskip 1em plus 0.5em minus 0.4em\relax Montreal, QC, Canada: IEEE, Oct. 2021, pp. 15\,213--15\,222. [Online]. Available: \url{https://ieeexplore.ieee.org/document/9709992/}
\BIBentrySTDinterwordspacing

\bibitem{yue2022nspsfm}
\BIBentryALTinterwordspacing
J.~Yue, D.~Manocha, and H.~Wang, ``\BIBforeignlanguage{en}{Human {Trajectory} {Prediction} via {Neural} {Social} {Physics}},'' in \emph{\BIBforeignlanguage{en}{Proceedings of the {European} {Conference} on {Computer} {Vision} ({ECCV})}}.\hskip 1em plus 0.5em minus 0.4em\relax arXiv, July 2022, arXiv:2207.10435 [cs]. [Online]. Available: \url{http://arxiv.org/abs/2207.10435}
\BIBentrySTDinterwordspacing

\bibitem{sun_move_2021}
\BIBentryALTinterwordspacing
M.~Sun, F.~Baldini, P.~Trautman, and T.~Murphey, ``Move {Beyond} {Trajectories}: {Distribution} {Space} {Coupling} for {Crowd} {Navigation},'' in \emph{Robotics: {Science} and {Systems} {XVII}}.\hskip 1em plus 0.5em minus 0.4em\relax Robotics: Science and Systems Foundation, July 2021. [Online]. Available: \url{doi.org/10.15607/RSS.2021.XVII.053}
\BIBentrySTDinterwordspacing

\bibitem{chen_relational_2020}
\BIBentryALTinterwordspacing
C.~Chen, S.~Hu, P.~Nikdel, G.~Mori, and M.~Savva, ``\BIBforeignlanguage{en}{Relational {Graph} {Learning} for {Crowd} {Navigation}},'' in \emph{\BIBforeignlanguage{en}{2020 {IEEE}/{RSJ} {International} {Conference} on {Intelligent} {Robots} and {Systems} ({IROS})}}.\hskip 1em plus 0.5em minus 0.4em\relax Las Vegas, NV, USA: IEEE, Oct. 2020, pp. 10\,007--10\,013. [Online]. Available: \url{https://ieeexplore.ieee.org/document/9340705/}
\BIBentrySTDinterwordspacing

\bibitem{Everett2021}
M.~Everett, Y.~F. Chen, and J.~P. How, ``{Collision avoidance in pedestrian-rich environments with deep reinforcement learning},'' \emph{IEEE Access}, vol.~9, pp. 10\,357--10\,377, 2021.

\bibitem{samavi2024sicnav}
\BIBentryALTinterwordspacing
S.~Samavi, J.~R. Han, F.~Shkurti, and A.~P. Schoellig, ``{SICNav: Safe and Interactive Crowd Navigation Using Model Predictive Control and Bilevel Optimization},'' \emph{IEEE Transactions on Robotics}, vol.~41, pp. 801--818, 2024. [Online]. Available: \url{http://sepehr.fyi/projects/sicnav}
\BIBentrySTDinterwordspacing

\bibitem{vandenBerg2011orca}
\BIBentryALTinterwordspacing
J.~van~den Berg, S.~J. Guy, M.~Lin, and D.~Manocha, ``\BIBforeignlanguage{en}{Reciprocal n-{Body} {Collision} {Avoidance}},'' in \emph{\BIBforeignlanguage{en}{Robotics {Research}}}, B.~Siciliano, O.~Khatib, F.~Groen, C.~Pradalier, R.~Siegwart, and G.~Hirzinger, Eds.\hskip 1em plus 0.5em minus 0.4em\relax Berlin, Heidelberg: Springer Berlin Heidelberg, 2011, vol.~70, pp. 3--19, series Title: Springer Tracts in Advanced Robotics. [Online]. Available: \url{http://link.springer.com/10.1007/978-3-642-19457-3_1}
\BIBentrySTDinterwordspacing

\bibitem{hess2016cartograper}
W.~Hess, D.~Kohler, H.~Rapp, and D.~Andor, ``Real-time loop closure in 2d lidar slam,'' in \emph{IEEE International Conference on Robotics and Automation (ICRA)}, 2016, pp. 1271--1278.

\bibitem{wang2024yolov9}
C.-Y. Wang and H.-Y.~M. Liao, ``Yolov9: Learning what you want to learn using programmable gradient information,'' 2024.

\bibitem{burnett2019autotrack}
\BIBentryALTinterwordspacing
K.~Burnett, S.~Samavi, S.~L. Waslander, T.~D. Barfoot, and A.~P. Schoellig, ``{aUToTrack} : {A} {Lightweight} {Object} {Detection} and {Tracking} {System} for the {SAE} {AutoDrive} {Challenge},'' in \emph{Conference on {Computer} and {Robot} {Vision} ({CRV})}, 2019. [Online]. Available: \url{https://ieeexplore.ieee.org/abstract/document/8781613}
\BIBentrySTDinterwordspacing

\bibitem{lee2022patchwork}
S.~Lee, H.~Lim, and H.~Myung, ``Patchwork++: Fast and robust ground segmentation solving partial under-segmentation using 3d point cloud,'' in \emph{2022 IEEE/RSJ International Conference on Intelligent Robots and Systems (IROS)}, 2022, pp. 13\,276--13\,283.

\bibitem{rosmann2015costmapconverter}
\BIBentryALTinterwordspacing
C.~Rosmann, ``costmap\_converter ros package,'' 2015. [Online]. Available: \url{https://wiki.ros.org/costmap_converter}
\BIBentrySTDinterwordspacing

\bibitem{verschueren2021acados}
\BIBentryALTinterwordspacing
R.~Verschueren, G.~Frison, D.~Kouzoupis, J.~Frey, N.~van Duijkeren, A.~Zanelli, B.~Novoselnik, T.~Albin, R.~Quirynen, and M.~Diehl, ``acados -- a modular open-source framework for fast embedded optimal control,'' \emph{Mathematical Programming Computation}, Oct 2021. [Online]. Available: \url{https://doi.org/10.1007/s12532-021-00208-8}
\BIBentrySTDinterwordspacing

\bibitem{samavi2025sicnavdiff}
S.~Samavi, A.~Lem, F.~Sato, S.~Chen, Q.~Gu, K.~Yano, A.~P. Schoellig, and F.~Shkurti, ``Sicnav-diffusion: Safe and interactive crowd navigation with diffusion trajectory predictions,'' \emph{arXiv preprint arXiv:2503.08858}, 2025.

\bibitem{lai2023spherical}
X.~Lai, Y.~Chen, F.~Lu, J.~Liu, and J.~Jia, ``Spherical transformer for lidar-based 3d recognition,'' in \emph{Proceedings of the IEEE/CVF Conference on Computer Vision and Pattern Recognition}, 2023, pp. 17\,545--17\,555.

\end{thebibliography}

\end{document}